\documentclass[10pt, conference]{ieeeconf}      

\IEEEoverridecommandlockouts                              

\usepackage[ruled,vlined]{algorithm2e}
\usepackage{url}

\newcommand{\qed}{\nobreak \ifvmode \relax \else
      \ifdim\lastskip<1.5em \hskip-\lastskip
      \hskip1.5em plus0em minus0.5em \fi \nobreak
      \vrule height0.75em width0.5em depth0.25em\fi}

\usepackage{amsmath,amssymb}

\usepackage{tabularx}
\usepackage{tikz,hyperref,graphicx,units}
\usepackage{subfigure}
\usepackage{benktools}
\usepackage{bbm}
\renewcommand{\baselinestretch}{.5}

\usepackage{caption}
\usepackage{epstopdf}

\usepackage{sidecap,wrapfig}
\usepackage[ruled,vlined]{algorithm2e}

\newcommand{\bu}{\mathbf{u}}

\newcommand{\bx}{\mathbf{x}}

\newcolumntype{L}[1]{>{\RaggedRight\hspace{0pt}}p{#1}}
\newcolumntype{R}[1]{>{\RaggedLeft\hspace{0pt}}p{#1}}

\newboolean{include-notes}
\setboolean{include-notes}{true}
\newcommand{\adnote}[1]{\ifthenelse{ \boolean{include-notes}}%
 {\textcolor{blue}{\textbf{#1}}}{}}
 
 \newcommand{\sknote}[1]{\ifthenelse{ \boolean{include-notes}}%
 {\textcolor{blue}{\textbf{SK: #1}}}{}}
 
  \newcommand{\mlnote}[1]{\ifthenelse{ \boolean{include-notes}}%
 {\textcolor{purple}{\textbf{ML: #1}}}{}}
 
 \newcommand{\jmnote}[1]{\ifthenelse{ \boolean{include-notes}}%
 {\textcolor{orange}{\textbf{JM: #1}}}{}}

\renewcommand{\baselinestretch}{0.95}
\usepackage{times}
\usepackage{microtype}

\title{ Learning Robust Bed Making \\
 using Deep Imitation Learning with DART
}

\author{Michael Laskey$^1$, Chris Powers$^1$, Ruta Joshi$^1$, Arshan Poursohi$^3$, Ken Goldberg$^{1,2}$
\thanks{$^1$ Department of Electrical Engineering and Computer Sciences; {\small \{mdlaskey, chris\_powers, rjoshi\}@berkeley.edu} }%
\thanks{$^2$ Department of Industrial Engineering and Operations Research; {\small goldberg@berkeley.edu}}%
\thanks{$^{1-2}$ University of California, Berkeley;  Berkeley, CA 94720, USA}
\thanks{$^{3}$ Toyota Research Institute}%
}

\begin{document}

\maketitle
\thispagestyle{empty}
\pagestyle{empty}



\begin{abstract}
Bed-making is a universal home task that can be challenging for senior citizens due to reaching motions. Automating bed-making has multiple technical challenges such as perception in an unstructured environments, deformable object manipulation, obstacle avoidance and sequential decision making. We explore how DART, an LfD algorithm for learning robust policies, can be applied to automating bed making without fiducial markers with a Toyota Human Support Robot (HSR).  By gathering human demonstrations for grasping the sheet and failure detection, we can learn deep neural network policies that leverage pre-trained YOLO features to automate the task. Experiments with a ½ scale bed and distractors placed on the bed, suggest policies learned on 50 demonstrations with DART achieve $96\%$ sheet coverage, which is over $200\%$ better than a corner detector baseline using contour detection.
 \end{abstract}

\section{Introduction} 
Home robotics offer the potential to provide treatment and care to senior citizens. A common home task is bed-making~\cite{clark1990older,fausset2011challenges}, which can be physically challenging due to the bending and leaning movements required. Additionally, senior citizens prefer a robot assistant over a human assistant for bed-making due to reduced intrusiveness in their private space~\cite{beer2012domesticated}. However, automating the task of making a bed present numerous technical challenges, such as perception in an unstructured environment, deformable object manipulation and sequential decision making.

To tackle these challenges, we formulate the bed making problem as optimizing the coverage of a bed sheet along a given bed frame. A solution to this objective is grasping the corners of each sheet and pulling them towards the end of the bed frame. However, location of the sheet corner is non-trivial because 1 it can be covered by the sheet during the bed making process and 2) in a home environment it is possible for people to leave arbitrary items on the bed such as stuffed animals, toys or clothing, which can confuse a vision system.

To address these difficulties, we propose learning from a human supervisor's demonstrations with Off-Policy Imitation Learning, where a robot observes a demonstration and learns a mapping from state to control via regression~\cite{dart}. Imitation Learning has been shown to be successful in domains such as self-driving cars~\cite{pomerleau1989alvinn}, quadcopter flight~\cite{ross2013learning} and grasping in clutter~\cite{laskeyrobot}. 

To perform the task, the learned agent, must select where to grasp the bed-sheet, a grasping policy, and whether or not the stretch was successful, a transition policy.  We chose to learn these components because they require making decision based on visual data. The rest of the system can be designed using the HSR's internal motion planner because it only involves motion of the body.

Imitation learning on visual information can require training data intensive neural networks. In order to learn the task using a small amount of data, we use a transfer learning algorithms that leverages a pre-trained object detection network, YOLO. Object detection is similar task to our supervisor's actions in that it both requires identifying a points on the scene and classification. Thus, a network trained on a very large dataset may contain relevant features for identifying these points. 

Applying Off-Policy Imitation Learning to sequential tasks can be difficult due to compounding error~\cite{ross2010reduction}. When the supervisor provides demonstrations it is unlikely for her to make a mistake, which means no example of how to recover are shown to the robot.  This mismatch between training and test distribution is known as the covariate shift. Thus, when the robot makes a mistake it will deviate from training data and not be able to recover. In our bed-making system this effect causes our transition policy to be biased towards stating the robot successfully grasped the sheet.

\begin{figure}
\center
\includegraphics{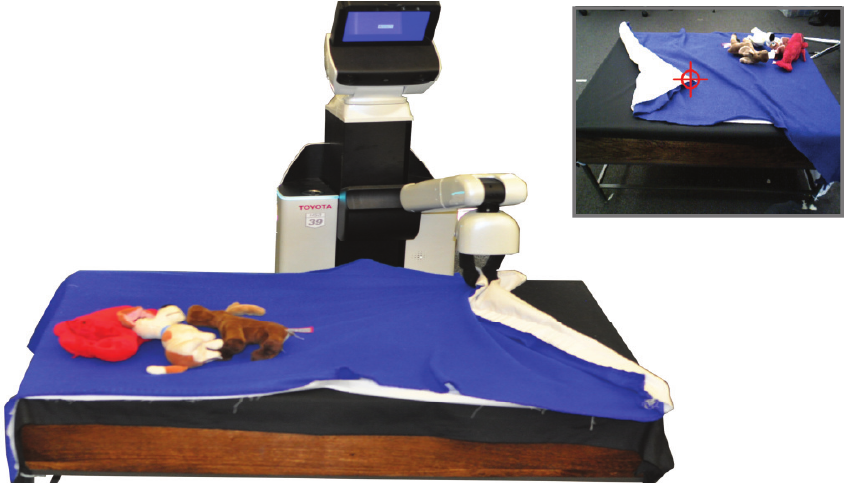}
\caption{
  \footnotesize
A Toyota Human Support Robot (HSR) stretching a sheet to make the bed, which is $\frac{1}{2}$ size of a twin bed. The robot was trained on 50 demonstrations of a human supervisor teaching it how to make a bed and is now being tested on a situation where other household objects, such as stuffed animals, are placed on the bed. In the top right hand corner, a prediction from the learned grasp selection is shown. The prediction still correctly grasps the corner of the bed sheet despite the additional objects placed on the bed.}
\vspace*{-20pt}
\label{fig:teaser}
\end{figure}

One way to correct for covariate shift is to roll out the current robot’s policy (i.e. allow it to visit error states) and provide corrective feedback~\cite{ross2010reduction} However, during training the robot can be highly sub-optimal and execution of the policy might result in unsafe collision with the surroundings. Recently, a new algorithm, DART, has been proposed to inject small optimized noise into the supervisor's policy to simulate the trained robot's error and show it how to recover.  DART maintains a high level of performance during the collection process because it only simulates the trained robot's policy, this is advantageous in bed-making because large error can result in collisions with the surrounding environment. 

This paper makes three contributions: 

\begin{enumerate}
\item The first formulation, to our knowledge, of the bed making problem. 
\item A robotic system architecture for robustly making a bed under external perturbations such as the placement of household objects on the bed and lighting. 
\item Experimental evaluation that compares DART to a corner detection heuristic and traditional Off-Policy learning (i.e. Behavior Cloning).
\end{enumerate}

We evaluate performance of our system with a Toyota HSR on a bed setup that is $\frac{1}{2}$ the size of a twin bed. We show when our neural network policy is trained with 50 demonstrations collected with DART can match the supervisor within $< 2\%$ of sheet coverage when no objects are added. When unseen household objects are placed on the bed, DART achieves $96\%$ sheet coverage, which is a $200\%$ increase over a corner detection baseline using contour detection.

\section{Related Work}
We survey the related work on cloth based manipulation and Imitation Learning.

\noindent \textbf{Cloth Based Manipulation}
Manipulation of cloth has been explored in a variety of contexts including laundry folding and surgical robotics. Shepard et al. used an algorithm based on identifying and tensioning corners to enable a home robot to fold laundry~\cite{maitin2010cloth}. Balaguer et al. proposed a technique that used 3rd person human demonstrations to learn how to fold a towel in isolation~\cite{balaguer2011combining}. Cusumano et al. examined the problem of bringing clothing into an arbitrary position and proposed using a deformable object simulator to plan motions~\cite{cusumano2011bringing}. Shibata proposed to fold a towel without using registration by examining humans performing the action and designing a robust folding strategy~\cite{shibata2012trajectory}, given the known initial position of the towel.

\begin{figure}
\center
\includegraphics{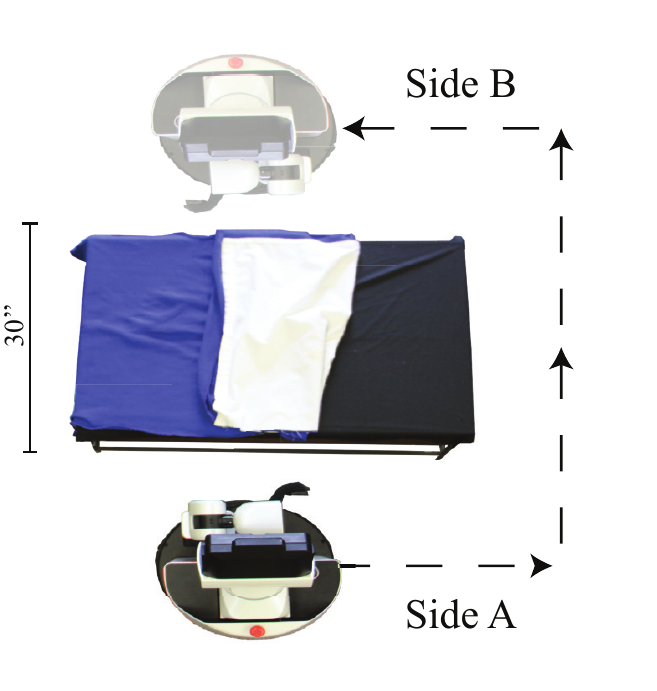}
\caption{
   \footnotesize
An overhead view of the bed making setup. The robot starts on Side A of the bed and executes the bed sheet stretching option until it believes the sheet is successfully stretched. At which point, the robot traverses along the indicated path marked with a dashed line to Side B using its mobile base.  Once, at Side B the robot again executes the sheet stretching option until it believes Side B is also successfully stretched. The robot then returns back to Side A and terminates the program.}
\vspace*{-20pt}
\label{fig:bed_top}
\end{figure}

In the surgical setting, cutting of cloth has been considered due to it being similar to cutting tissue. Murali et al. examined cutting a circle out of surgical gauze via leveraging expert demonstrations~\cite{murali2015learning}. However, this approach suffered in reliability due to imprecision in the tensioning policy on the cloth. Thanajeyan et al. examined learning a more robust tensioning policy in simulation using state of the art Deep RL algorithms~\cite{thananjeyan2017multilateral}.

As opposed to these work, Bed-Making requires detection of grasp points in an unstructured environment that is subject to change.

\noindent \textbf{Imitation Learning}
Imitation Learning has been shown to be successful in domains such as self-driving cars~\cite{pomerleau1989alvinn}, quadcopter flight~\cite{ross2013learning} and grasping in clutter~\cite{laskeyrobot}. In general, Imitation Learning algorithms are either off-policy or on-policy. In off-policy Imitation Learning, the robot passively observes the supervisor, and learns a policy mapping states to controls by approximating the supervisor's policy. This technique has been successful, for instance, in learning visuomotor control policies for self-driving cars~\cite{pomerleau1989alvinn,bojarski2016end}.

 However, when applying this technique Pomerleau et al. observed that the self-driving car would steer towards the edge of the road during execution and not be able to recover~\cite{pomerleau1989alvinn}. Ross and Bangell theoretically showed that this was due to the robot's distribution being different than the supervisor's, a property known as covariate shift, which caused errors to compound during execution~\cite{ross2010efficient}.

Ross et al.~\cite{ross2010reduction} proposed DAgger, an on-policy method in which the supervisor iteratively provides corrective feedback on the robot's behavior. This alleviates the problem of compounding errors, since the robot is trained to identify and fix small errors after they occur. However, this can be problematic for bed-making because the robot needs to physically execute potentially highly sub-optimal actions, which may lead to collisions. Recently, it was shown that another way to correct for covariate shift is to inject small noise levels into the supervisor’s policy to simulate error during data collection~\cite{dart}. A technique, known as DART, was proposed to optimize for this noise distribution. We demonstrate that by using DART, we can achieve robust bed making and collect data at the same level of performance as the supervisor.

\begin{figure*}
 \center
\includegraphics{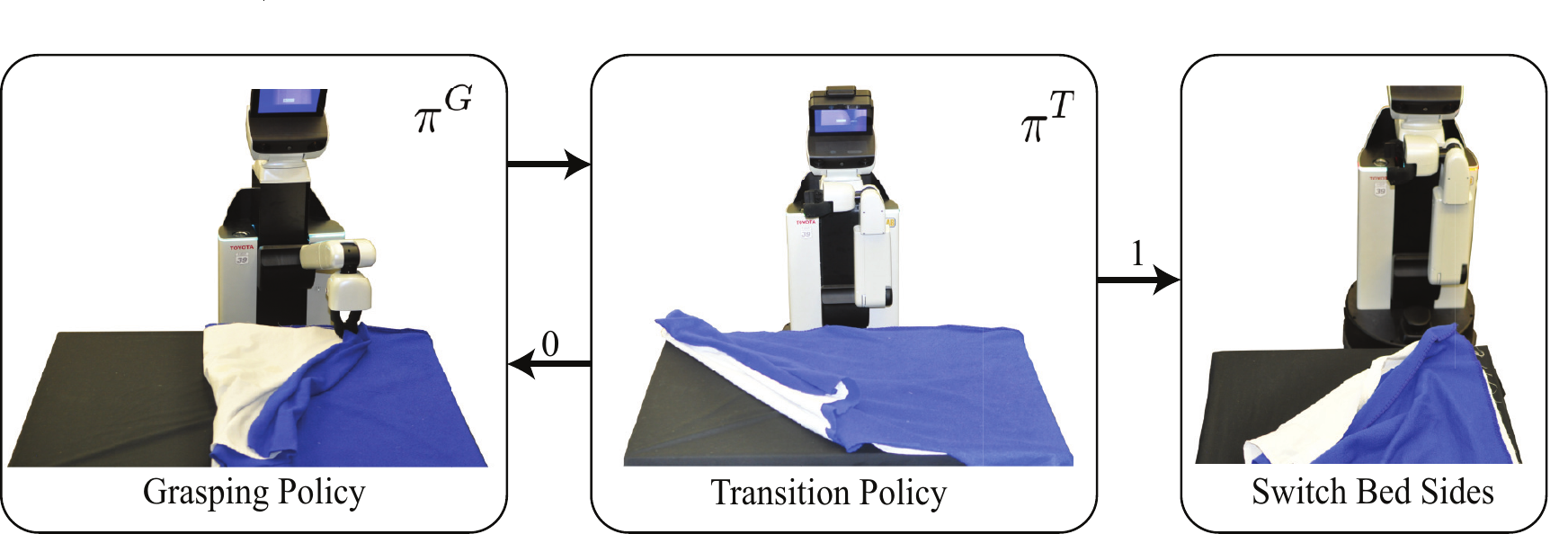}
\caption{
    \footnotesize
The two policies learned to make a bed composed as an option. The grasping policy, $\pi^G$, selects where a grasp should be executed. Given a grasp the robot then uses its motion planner to stretch the sheet towards the top of the bed. Afterwards, the transition policy, $\pi^T$, decides whether or not the stretching motion was successful (i.e. did the corner of the sheet reach the end of the bed frame). If it is successful (i.e $\pi^T(x) = 1$) then the robot switches to the other side of the bed. If it is not successful (i.e $\pi^T(x) = 0$), the robot re-attempts the grasp and stretch motion.}
\label{fig:bed_options}
\vspace*{-15pt}
\end{figure*}

\section{Problem Statement and Background}\label{sec:PS}
\noindent \textbf{Assumptions} We assume that the robot has the ability to reach the surface of the bed-frame at any point and is able to register the bed into its coordinate space. We additionally assume that the bed sheet is placed initially such that both of the corners are visible from the camera.

\subsection{Definitions}

\noindent \textbf{Robot} A mobile manipulator robot can be referenced by 2 different coordinate frames; camera and gripper. We will denote a 6D pose, $T$ as parameterized by $R$ and $t$, which represent the rotation and translation parameters. The robot's head mounted camera's pose will be denoted $T_C$.  The robot's gripper pose will be represented by $T_G$.

\noindent \textbf{Bed} A bed is composed of a bed frame and a bed sheet. The bed frame is a rigid rectangular structure whose 6D pose is given by $T_P$ with dimensions $W_F \times H_F \times L_F$. The physical space occupied by the bed can be defined formally via an occupancy function $B : \mathbb{R}^3 \rightarrow \lbrace 0, 1 \rbrace$, which determines if a point in 3D space is part of the bed-frame or not.  The sheets are deformable cloth with dimension $W_S \times H_S \times L_S$, which is also represented with an occupancy function $\xi:\mathbb{R}^3 \rightarrow \lbrace 0, 1 \rbrace$ (i.e. $W_\xi = W_B$ and $L_\xi = L_B$). The complete state of a bed can be described by the set $ \lbrace{B, \xi \rbrace}$. Our bed setup is shown in Fig. \ref{fig:teaser}. Throughout the paper we will refer to two sides of the bed that the robot will be on: Side A and Side B. These are illustrated in Fig. \ref{fig:bed_top}.

\subsection{Objective}
We write the objective of bed making, in terms of surface coverage of the sheets over the bed frame. Thus, the goal of the robot is to stretch the sheet over the bed frame. We formalize this as the following objective:

\begin{equation}\label{eq:main_obj}
\underset{\xi}{\mbox{max}} \: \int_x \mathbbm{1} (B(x) + \xi (x) = 2) dx
\end{equation}

which corresponds to increasing the overlap between bed-frame and sheets, or the coverage.

One way to solve Eq. \ref{eq:main_obj}, requires increasing the coverage of the cloth, which can be solved by grasping at the corners of the sheet and stretching them towards the bed frame. However, this approach can suffer from several challenges.  

First, even if the sheet initially has both corners exposed to the camera during stretching of the cloth the other corner maybe become folded over, as shown in Fig. \ref{fig:dart_rollout}. Second, corner extraction can be hard when the environment is subject to changes in lighting and additional objects are added to the scene, such as objects being placed on the bed. Inspired by recent successes in deep imitation learning, we propose to learn the stretching policy from demonstrations. 

\subsection{Learning Bed Making Options}
To optimize Eq. \ref{eq:main_obj}, we want to learn where to grasp the sheet to execute a stretching motion. To be robust to failure, we also want to learn to detect for whether the robot should retry stretching the bed sheet or transition to the other side of the bed. Learning these two components, the grasping policy and the transition policy can be formulated as an \emph{option}~\cite{fox2017multi}. An illustration of our bed making option can be seen in Fig.  \ref{fig:bed_options}. 

Denote the grasping policy $\pi^G  : \mathbb{R}^{640\times480\times3} \rightarrow \mathbb{R}^{2}$, or a mapping of RGB images from the head mounted Primensense to a pixel position of where to grasp. In Sec. \ref{sec:grasp} , we describe how to turn this pixel position into a 6 DOF gripper pose. The transition policy is represented as $\pi^T: \mathbb{R}^{640\times480\times3} \rightarrow \lbrace 0, 1\rbrace$, which maps images of the workspace to a binary decision of whether to transition or not.

In Imitation Learning, a supervisor is a policy which given a state can provide the correct control for the robot to take. The supervisor's policies for the transition and grasping policies are denoted by $\tilde{\pi}^G$ and $\tilde{\pi}^T$, respectively. Supervision comes from a human operator, who given an image returns a 2D grasp point or a binary decision depending on which policy is being queried.  The goal is to learn a representation for these policies with a neural network, which will be denoted $\pi^G_\theta$ and $\pi^T_\theta$, where $\theta$ are the parameterized weights learned by the network.

In the options framework~\cite{fox2017multi}, at each state the robot decides on a grasp and executes a stretching motion. The robot then queries the transition policy for whether or not it should retry. We can denote these two policies together as the composition, $\pi = \pi^G \bowtie  \pi^T$, where $\pi: \mathcal{X} \rightarrow \mathcal{U}$. $\mathcal{X}$ denotes the state space of images before and after the grasping policy is executed (i.e. $\mathbb{R}^{640x480x3} \times \mathbb{R}^{640x480x3} \in \mathcal{X}$). $\mathcal{U}$  is the composition of grasp location and the decision whether or not to transition (i.e. $\mathbb{R}^2 \times \lbrace 0, 1 \rbrace \in \mathcal{U}$). The option policy induces the following Markov Chain over a trajectory $\tau$:

$$p(\tau|\pi) = p(\bx_0) \prod^\infty_{t=0} p(\bx_{t+1}|\bx_t,\bu_t) p(\bu_t|\pi,\bx_t)$$,

 which has an infinite time horizon because the policy will eventually choose to stay in a termination state forever. 

In Imitation Learning, we want to learn a policy, $\pi_{\theta}$ that matches the supervisor, $\tilde{\pi}$.  We measure how close two policies are by a function known as a surrogate loss, $l : \mathcal{U} \times \mathcal{U} \rightarrow \mathbb R^+$. We specifically consider the following surrogate loss: 

\begin{align*}
&l(\tilde{\pi}(\bx),\pi_{\theta}(\bx)) = \\
&|| \tilde{\pi}^G(\bx_1) - \pi_{\theta}^G(\bx_1)||_2 + |\tilde{\pi}^T(\bx_2) - \pi_{\theta}^T(\bx_2)|
\end{align*}

, which is composed of a Euclidean Distance for the grasping policy and a binary loss for the transition policy. Note $\bx_1$ and $\bx_2$ correspond to the two images before and after execution of the stretching motion and together create the composition state $\bx$.

Given a surrogate loss function, we want to find a policy that minimizes the expected surrogate loss under the distribution of states the robot is likely to visit: 

\begin{equation}\label{eq:sur_loss}
\underset{\theta}{\mbox{min}} \: E_{p(\tau|\pi_{\theta})} \sum^{\infty}_{t=0} l(\tilde{\pi}(\bx_t),\pi_{\theta}(\bx_t))
\end{equation}

Unlike traditional supervised learning, optimization of Eq. \ref{eq:sur_loss} is challenging because the distribution of states trained on depends on $\theta$. Thus, in practice it is common to sample demonstrations from another distribution, such as the supervisor's $p(\tau|\tilde{\pi})$ and minimize the sample surrogate loss:

\begin{equation}\label{eq:sur_loss}
\underset{\theta}{\mbox{min}} \: E_{p(\tau|\tilde{\pi})} \sum^{\infty}_{t=0} l(\tilde{\pi}(\bx),\pi_{\theta}(\bx)),
\end{equation}

training on the  supervisor's distribution is an approached commonly known as Behavior Cloning.

Behavior Cloning though has been shown to suffer from an issue known as covariate shift~\cite{ross2010efficient}. Covariate shift occurs because the data collected on the supervisor's distribution $p(\tau|\tilde{\pi})$ may not be reflective of the data likely on the robot's distribution $p(\tau|\pi_{\theta})$.

\subsection{Reducing Covariate Shift with DART}
One way to correct for covariate shift is to inject small levels of noise into the supervisor's policy to simulate errors that are likely to occur under the robot's policy. DART is a recent algorithm that formulates this as choosing the noise term that maximizes the likelihood of applying the trained robot's controls. 

In the context, of the options setting we chose to only inject noise into the grasping policy because we predict the transition policy will have small error. We inject Gaussian noise in the supervisor's grasping policy, $\tilde{\pi}^G$. Specifically, the probability of applying the grasp point $\bu_1$ is $\mathcal{N}(\bu_1|\tilde{\pi}^G(\bx_1),\Sigma)$, where the mean is the supervisor's action $\tilde{\pi}^G(\bx_1)$ and the covariance matrix is  $\Sigma$. 

DART recommends computing $\Sigma$ by first training a policy on a smaller dataset and then maximizing the likelihood over a held out set of $K$ demonstrations.

$$\Sigma = \frac{1}{K} \sum^{K}_{k=0} (\pi_{\theta}^G(\bx_k) - \tilde{\pi}^G(\bx_k))(\pi_{\theta}^G(\bx_k) - \tilde{\pi}^G(\bx_k))^T$$

In this work, we first collect $10$ demonstrations with no noise injected and then compute $\Sigma$ using 4-fold cross validation. DART additionally recommends using a prior over the magnitude of the robot's final error to help scale the noise injected. From previous experiments we know the Euclidean loss achieved by the robot on this task is around 30 pixels of error. Thus, we scale the $\Sigma$ by this quantity using the technique prescribed by DART. 

\begin{figure}
\center
\includegraphics{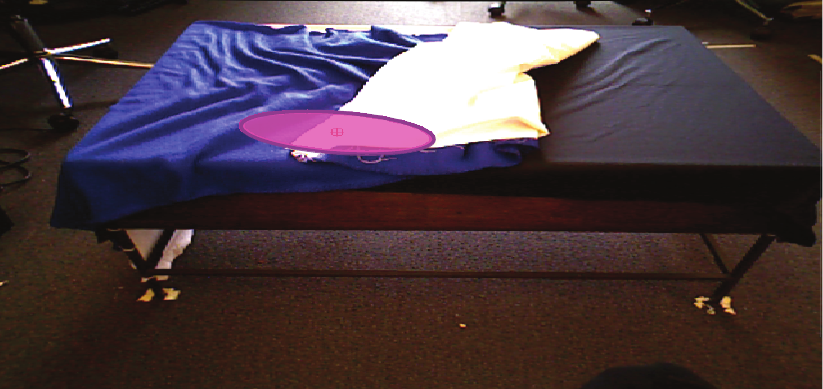}
\caption{
   \footnotesize
An illustration of the Gaussian Noise injected into the supervisor's grasp policy that is computed with DART. The pink region represents the area of points that are selected with $95\%$ probability. The Gaussian noise is larger in the horizontal axis, because the robot has more difficulty predicting this direction. This difficulty could arise from the fact that the sheet corner varies more in the horizontal axis than the vertical axis during training. 
 }
\vspace*{-20pt}
\label{fig:dart_noise}
\end{figure}

The Gaussian noise injected by DART is shown in Fig. \ref{fig:dart_noise}. The noise is spread out more in the horizontal direction indicated the trained robot has larger error in this direction. Intuitively, this makes sense because the bed sheet corner varies more along this axis.

 \section{Experimental System}

 \subsection{Robot and Bed Testbed}
Our system uses a Toyota HSR robot. The HSR is a mobile home robot that has 7 DOF, with 4 DOF in its upper-body arm and 3 DOF in the mobile base. The HSR comes equipped with a built in motion planner that plans a trajectory to a given goal for both its arm and mobile base. The accuracy of the HSR when commanded to move to a target position is on the order of 2 centimeters in translational pose. 

The HSR additionally has ability to track where it is with respect to a fixed world frame by using on board IMU sensors and wheel encoders in its base. The world frame is created every time the robot is turned on. Finally, the HSR performs on board sensor fusion to create a dynamic obstacle map for its mobile base using its LIDAR scanner. If an obstacle, such as a person, enters its path it will wait until they move. 

Our setup used a miniaturized bed frame, which is roughly $\frac{1}{2}$ the size of twin bed,  with dimensions $W_B = 30"$, $L_B = 40"$ and $H_B= 35"$. The bed consists of one sheet, which is cut to match the dimensions of the bed frame. The sheet is solid blue on top and solid white on the bottom. The bottom side of the sheet is fixed to the base of the bed frame to simulate the corners being tucked under the mattress, which is common for bed setups. The robot can register the bed to its world frame via an AR marker placed at the base of the bed on Side A. Using the HSR's stereo camera and known measurements of the bed, we can then compute where the 3D structure of the bed is. 

Lastly, the bed frame is placed next to a large window (on Side B), which causes changes in lighting throughout the day.  Additionally, people were allowed to freely walk in the robot's field of view and potentially place objects such as backpacks or lab equipment on the floor. Thus, the robot had to learn to be robust to these natural disturbances.  Example images from the robot camera can be seen in Fig. \ref{fig:dart_rollout}.

\subsection{Grasping Policy}\label{sec:grasp}
Given an image the grasping policy must select a 2D pixel location for the where the HSR should grasp the sheet. We can project $\bu_1$ onto the scene by first measuring the depth value, $z$, from the corresponding depth image taken from the Primesense Camera. 

\begin{figure}
\center
\includegraphics{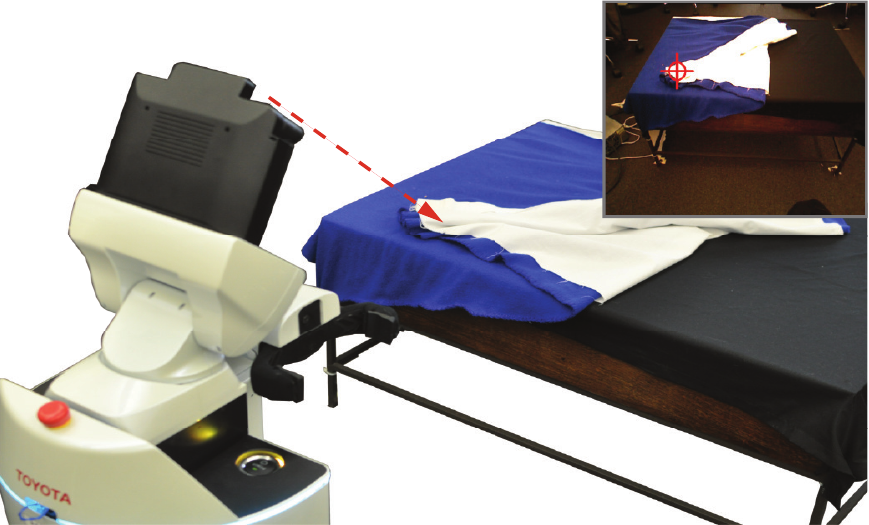}
\caption{
   \footnotesize
The interface used to select grasp point on the bed sheet. In the top right a selected grasp point is marked with a red cross hair. The HSR robot then projects this point onto the bed to determine where to place its gripper.}
\vspace*{-20pt}
\label{fig:gripper_sys}
\end{figure}

The $z$ value is determined via the median value of the points in a 10x10 bounding box centered around $\bu_1$. When computing the median value, we remove all points that correspond to missing data. The pixels and corresponding depth value can be combined to form $\mathbf{u}_c = [u_x,u_y,z]$. We can then compute the 3D point in robot frame, $\mathbf{u}_r$ via camera deprojection using the known camera parameters for the Primesense~\cite{horn1986robot}. 

$\mathbf{u}_r$ provides the 3D point in world frame, however it does not specify the rotation of the gripper. Since, we know the position of the bed frame $T_B$, we can rotate the gripper to be orthogonal to the table. In Fig. \ref{fig:gripper_sys}, we show an illustration of how $
\bu_1$ point is projected onto robot's workspace from an image.

\subsection{Sheet Stretching}
Once the grasp location has been determined, the robot moves its gripper to the location and closes. The robot then pulls the sheet towards the closest upper corner of the bed. Given that the sheet is the same size of the bed frame, it is possible for the robot to grasp a part that cannot be stretched to the top corner. Pulling on a stretched sheet can result in high force on the robot's wrist, which will cause the HSR's internal controller to shut down. 

To prevent shutting down, the robot uses its 3-axis force sensor in the wrist to measure the force exerted by the sheet. If the force in the y-direction becomes higher than $20$N, the robot releases the sheet from its gripper. Due to how the system is implemented the robot performs the stretching motion in 6 steps, where after each step the force reading is queried.

\subsection{Transition Policy}

\begin{figure}
\center
\includegraphics{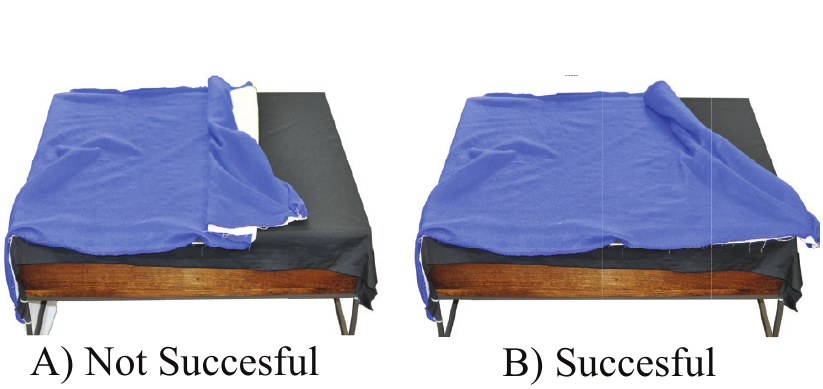}
\caption{
   \footnotesize
Examples of unsuccessful and successful grasp that the transition policy would need to classify. A) The stretching motion was unsuccessful because the sheet fell out of the gripper during the motion. B) The sheet is correctly stretched to the corner of the bed.   }
\vspace*{-20pt}
\label{fig:suc_state}
\end{figure}

After the robot attempts to stretch the sheet, it then checks whether the motion was successful. Due to imprecision in the learned grasp selection, it is possible that the robot fails to grasp the sheet in the correct area and needs to perform a re-grasp. Examples of failed and successful grasp can be in seen in Fig. \ref{fig:suc_state}. 

To decide whether or not the robot should retry it queries the transition policy $\pi_T$ and receives a binary signal $u_t$.  Once, the robot receives the signal that the bed sheet is fully stretched on one side, it uses its mobile base to follow a predefined path to switch to the other bed side.

 \begin{figure*}
 \center
\includegraphics{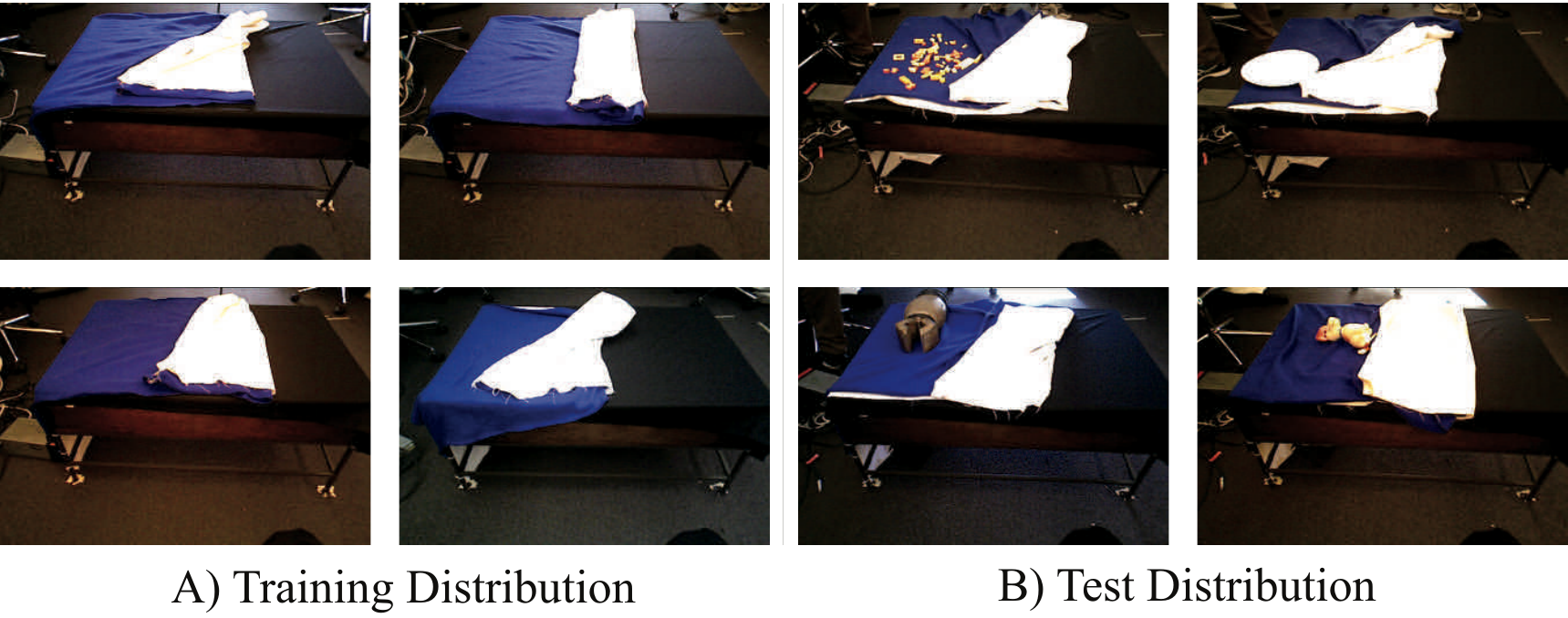}
\caption{
    \footnotesize
Examples of initial states shown to the robot, which are taken from the HSR’s head mounted primesense camera. The training distribution are states generated from our initial state sampler, which generates lines along the bed for the corners of the sheet to be aligned with. The test distribution also is generated with the same line sampler, however we additionally add household distractor objects such as legos, paper plates and stuffed animals. Notice that in both images the lighting in the room changes significantly and people are allowed to walk freely around the workspace. The robot must learn to be robust to these natural changes.   }
\vspace*{-15pt}
\label{fig:init_state}
\end{figure*}

\subsection{Neural Network Policies}
 For representing the policies $\pi^G_\theta$ and $\pi^T_\theta$, we trained two different neural networks. The networks trained are based on  YOLO~\cite{redmon2016you}, a state of the art real time object detection network. We modified the final layer of YOLO to only output a 2 dimensional prediction. We chose the YOLO architecture given the similarity of our task to object detection, which requires specifying a bounding box in 2D space and providing a classification label on the box. Additionally, YOLO has the advantages of being relatively faster compared to other object detection methods, which is ideal for real time control tasks~\cite{redmon2016you}.

For the grasping network, $\pi^G_{\theta}$, we use a squared Euclidean norm on the pixel wise distance between the predicted grasp point and the provided label. During training we scale the grasp labels to be zero mean and in the range of $[-1,1]$, to better condition the optimization. For the transition network $\pi^T_{\theta}$, we use a binary soft-max loss function, which returns the probability of whether the robot should transition or not. 

When re-training large network architectures such as YOLO, it is possible to overfit to smaller dataset being trained on. In light of this, we want to leverage pre-existing weights from a network trained on a larger dataset. By fine-tuning these existing weights on our dataset we hope to generalize better. The underlying hypothesize with this approach is that informative features are learned in the beginning convolutional layers, which can help reduce the state space of the problem.

Fine-tuning large networks like YOLO can have several challenges: 1) the optimization may move the weights far from the original features causing over fitting and 2) back propagation through a large number of convolutional layers can be computationally expensive. In light of this, we chose to fixe the first 26 convolutional layers and pre-compute these features before training. Razavian et al. showed on vision datasets that this can improve generalization during transfer learning~\cite{sharif2014cnn}.

In Table 1, we test the transfer learning approach by training on a dataset of 100 grasp examples collected with the Behavior Cloning policy. We compare no initialization of weights, fine-tuning the weights and fixing the first 26 convolutional layers. We train each network, on a Tesla K40 GPU, for 500 iterations and examine the test error on a held out dataset of 20 grasp examples. Interestingly, pre-computing the features leads to both better generalization and significantly faster training time. Thus, suggesting with limited data it is preferable to have less expressive models.

Finally, we apply data augmentation techniques to help increase the effective size of our data and expose the trained networks to more variations. We double our dataset by reflecting the image around the vertical axis. Intuitively, this transformation helps because a vertical flip creates an image that has the perspective similar to the opposite side of the bed. We also apply changes to the brightness and color of the image, which creates 6 more images. In total these augmentation techniques creates 12x more images.

\begin{table}[]
\centering
\begin{tabular}{l|l|l|l|}
\cline{2-4}
                                & Random Init. & Fine Tuned & Fixed Layers \\ \hline
\multicolumn{1}{|l|}{Test Loss (L2 Pixels)} &  293               & 111          & \textbf{41}           \\ \hline
\multicolumn{1}{|l|}{Time (m)}  & 31             & 31          &  \textbf{2} \\ \hline
\end{tabular}
\caption{
    \footnotesize
Different techniques for re-training the YOLO architecture. Random Initialization corresponds to not using pre-trained feature. Fine tuning loads pre-trained YOLO features into the architecture and then optimizes. Fixed Layer precomputes features from the first 26 convolutional layers and only optimizes the last layer. The more biased Fixed Layer generalizes better on the limited dataset and has the advantage of training significantly faster. }
\vspace*{-20pt}
\end{table}

\section{Experiments}

Our experiments are designed to test how well out learned polices perform on the distribution of initial states they are trained on and how well they generalize to hard unseen states with novel objects placed on the bed. 

 To standardize the distribution, the robot is trained on, we made a protocol for sampling initial states. Random positions for the two bed sheet corners were generated by selecting 2d coordinates from a uniform distribution with its origin at the center of the bed in the robot's coordinate frame. 
 
 The corner closest to the robot was sampled from a uniform distribution over the bottom left quarter of the bed, from the robot's visual perspective. The corner further from the robot was sampled from a uniform distribution over the top right quarter of the bed, from the robot's visual perspective. A red line segment was drawn between these two points on a user interface so that the supervisor performing the demonstration or running the experiment could set the edge of the sheet along the line. Examples of states drawn from this distribution can be seen in Fig. \ref{fig:init_state}.

\subsection{System Evaluation}

 \begin{figure*}
 \center
\includegraphics{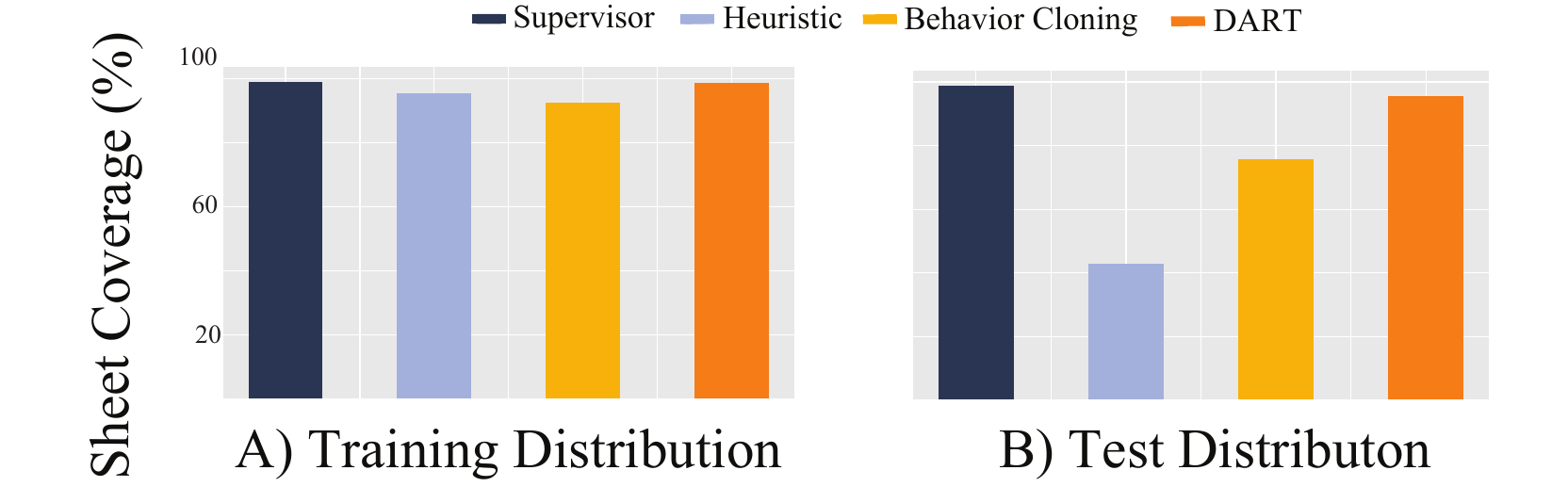}
\caption{
    \footnotesize
The  percentage of sheet coverage after the robot transitioned out of Side B. We compare 4 different methods; the human supervisor, a heuristic based on contour detection,  Behavior Cloning and DART. We report result averaged over 10 trials for states sampled from the initial state distribution, which we denote as the Training Distribution and for those with the Household Distractor Object (i.e. Test Distribution). On the Training Distribution all methods have similar performance to the Supervisor, however on the Test Distribution the policies trained with DART achieve 95\% sheet coverage, while Behavior Cloning achieves only 76\%.   }
\label{fig:sup_reward}
\vspace*{-20pt}
\end{figure*}
We evaluate how well the robot did in optimizing Eq. \ref{eq:main_obj}, by measuring the coverage of the sheet over the bedframe. The coverage is computed by first having a human provide the crop points for the bed frame and then using opencv contour detection to detect the sheet. Due to the changes in lighting, the human also needs to manually tune the color threshold periodically. The percentage area of the sheet that covers the image is then computed. We consider the final sheet coverage after the robot decides it is done stretching on Side B.  

\begin{figure}
 \center
\includegraphics{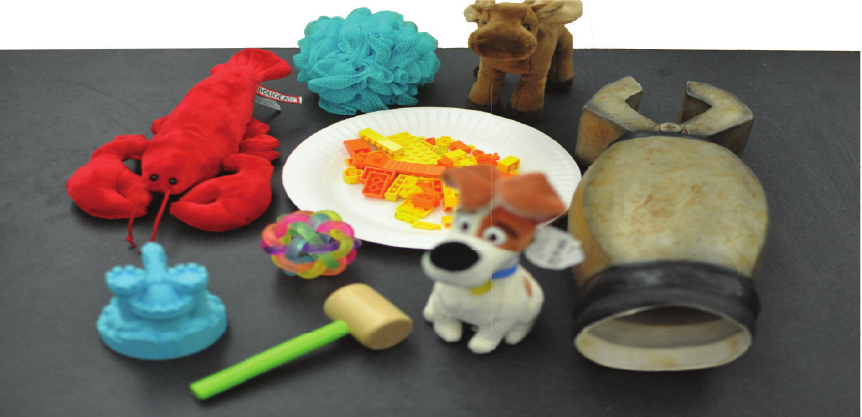}
\caption{
    \footnotesize
To test how robust our bed making system is, we place common household distractors object on top of the bed sheet. The grasping and transition policies were never exposed to these types of examples during training and need to be able to generalize. The distractors are also chosen due to their light weight (i.e. under 250g), which means they have a minimal effect on the dynamics of the bed making process, but only disrupt the vision system.   }
\label{fig:adv_object}
\vspace*{-15pt}
\end{figure}

We test each policies 10 times on two different scenarios. The first is on states sampled from the same initial state distribution that the policy was trained on, which we denote as the Training Distribution.  This scenario test how well the policy has learned to perform the task on states similar to what it saw in training. The second scenario tests how robust the policy is to changes in the environment. In order to measure how robust our learned policy is, we place household objects on the bed which have not been seen in training. The objects that we chosen to have varying color and geometry, for example a bright red lobster or a pile of legos. The objects are shown in Fig. \ref{fig:adv_object}. Examples of the initial states in both of these scenarios can be seen in Fig. \ref{fig:init_state}.

The first policy we evaluate is the supervisor's policy, which measures how good the robot can make the bed when a human is selecting the grasp point. We next evaluate the learned policies trained on 50 demonstrations that is either collected with Behavior Cloning (i.e. no noise injected) or DART. The policies are represented as the best neural network found in the previous section, which is the YOLO architecture with the first 26 convolutional layers fixed.

We finally compare the learned method to a heuristic approach. The heuristic uses contour detection and the bed sheet color to execute the strategy of always pulling the white left most pixel when the robot is on Side A and the blue right most pixel when the robot is on Side B. Given the initial states sample from our training distribution this could be a sensible strategy, since in general the bed sheet corner is white side up on Side A and blue side up on Side B.

In Fig. \ref{fig:sup_reward}, we report the percentage of sheet coverage when the robot decided to transition to finish the task. For similar states to the training distribution, we see all methods can achieve very high sheet coverage. The heuristic, Behavior Cloning and DART achieve 95\%, 92\% and 98\% coverage. Thus, suggesting it is possible for all methods to work reliably on states similar to training. 

\begin{figure}
 \center
\includegraphics{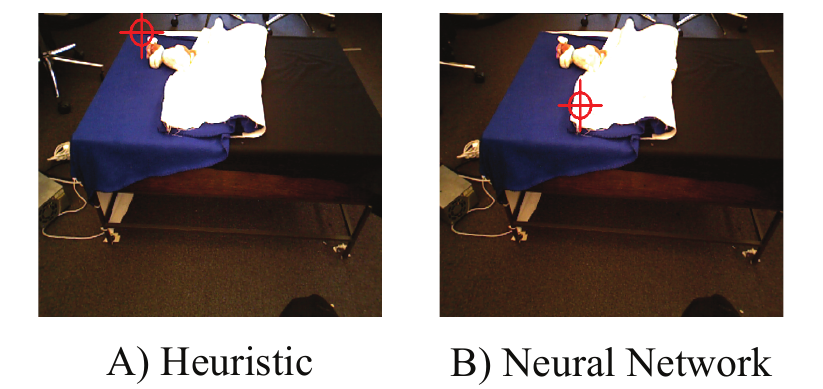}
\caption{
    \footnotesize
A sample state from the test distribution, where a stuffed white dog is placed on the bed. The heuristic method which uses contour detection to find the corner point is confused by the additional white object and selects a grasp point that is far away from the bed sheet corner. However, the learned network is not significantly affected by this and chooses a grasp point near the corner of the sheet.   }
\label{fig:anal_sucks}
\vspace*{-15pt}
\end{figure}

\begin{figure*}
 \center
\includegraphics{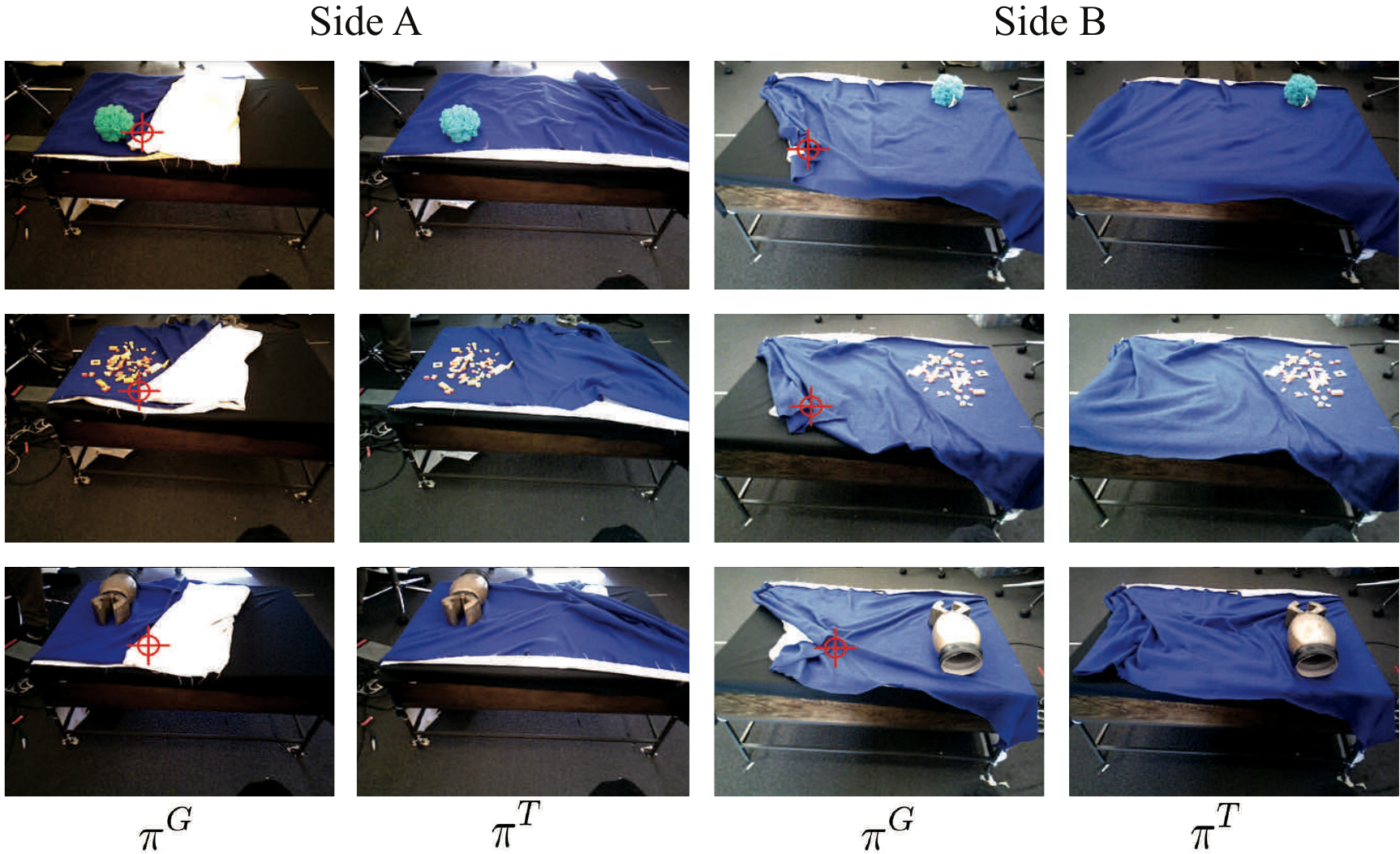}
\caption{
    \footnotesize
Example roll outs of the DART policy on the test distribution with the loofa, Legos and toy robot gripper placed on the bed (Top to Bottom). From Left to Right is the policy being rolled out, with the red cross hairs denoting where the grasp point is selected. The transition conditions policy always indicate success. The example on the bottom, with the toy gripper,  is an example of when complete coverage is not achieved. }
\label{fig:dart_rollout}
\vspace*{-20pt}
\end{figure*}

When the methods are evaluated on the test distribution, Behavior Cloning and the heuristic achieve 78\% and 49\% sheet coverage. One reason the heuristic approach performed so poorly is that objects with white texture would be mistaken for the sheet and cause the robot to collide with the bedframe. In Fig. \ref{fig:anal_sucks}, we show an example where the white stuff dog confuses the heuristic, but not the neural network.  DART maintains a robust performance and achieves 95\% sheet coverage. In Fig. \ref{fig:dart_rollout}, we show sampled roll outs of the policy, where the policy is robust to the distractor objects.

We finally, examine the time for the DART trained policy to complete the task. When averaged over 10 trials it took 3 minutes and 30 seconds to complete the task. With the evaluation of $\pi_{\theta}^G$ and $\pi_{\theta}^T$ taking a total of 0.5 seconds. The majority of the time can be attributed to the stretching motion and switching bide sides.

\subsection{Understanding DART's Performance}
\begin{figure}
 \center
\includegraphics{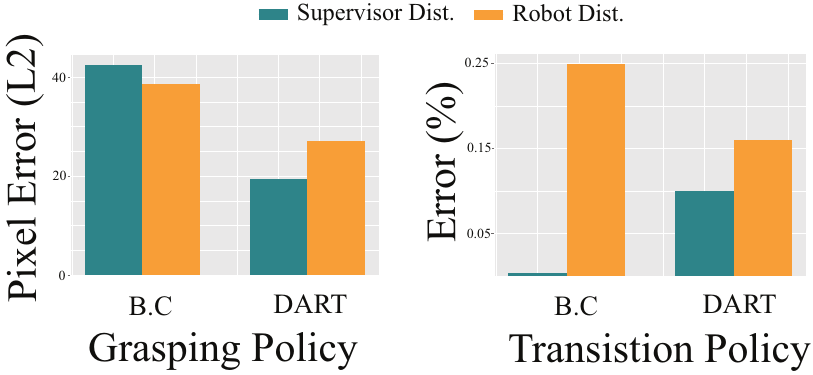}
\caption{
    \footnotesize
The error of the grasping policy and transition policy with respect to how well they matched the supervisor. The error of policies trained with DART and Behavior Cloning are both reported on data collected from the supervisor and on states visited when executing the policy. For both the grasping and transition policy, DART is able to achieve lower error because it caused to the supervisor to show the robot a larger diversity of states. 
  }
\label{fig:cs_graph}
\vspace*{-15pt}
\end{figure}

To better understand why DART achieved a gain in performance, we can measure the surrogate loss of the transition and grasping policy.  In Fig. \ref{fig:cs_graph}, we report the loss for each policy trained with Behavior Cloning and DART. To measure covariate shift, we evaluate the loss on on held out demonstrations from the supervisor and also during execution of the robot's policy. 

We see that DART's policies had lower loss when in both the grasping and transition policy,  $\pi^G$ and $\pi^T$, when deployed on the robot.  We attribute the lower error in the grasping policy to the fact that DART was shown a wider diversity of states during training, which allowed it to better generalize to unseen data. 

The transition policy is interesting because the error on the supervisor's distribution is very low for the Behavior Cloning policy, but much higher on the robot's distribution. This mismatch between the error on the supervisor's and robot's distribution can be attributed to covariate shift. Using Behavior Cloning, the supervisor only made $2$ failures during data collection, which meant the transition policy observed a very small set of failure modes. When DART was applied the supervisor made $27$ failures during data collection, which created a much more representative data-set. 

While DART caused more failures mode to occur, it is important to note that the supervisor with noise injected was still able to achieve $98\%$ sheet coverage during data collection. This high performance level was due to the fact that the errors were able to be recovered from because they were small. One downside of using DART is that each demonstration required on average 30 seconds more than with Behavior Cloning because the supervisor had to re-attempt a demonstration.

\section{Future Work and Conclusion}
We presented a technique for a mobile home robot to make a bed and be robust to common household items being placed on it. In order to achieve this robustness, we use recent advances in Off-Policy Imitation Learning to simulate small levels of error as the supervisor is collecting data. 

While, we are able to make a bed with our technique. There are several limitations that need to be addressed in future work. First, sometimes the sheet covers the bed frame, but has a lot of wrinkles in the sheet, this qualitatively leads to beds that do not look “made”. We hope to add an additional step to have the robot remove the wrinkles. Additionally, we assume the bed sheet initial states is with both corners of the sheet facing upwards. Examining ways to relax this constraint will be an exciting avenue for future work. For more information and videos see \url{https://people.eecs.berkeley.edu/~laskeymd/bed_making.html}.

\bibliography{references}

\end{document}